\documentclass[letterpaper, 10pt, conference]{ieeeconf}      % Use this line for a4 paper

\IEEEoverridecommandlockouts                           

\usepackage{amsmath,amssymb,amsfonts}
\usepackage{enumerate}
\usepackage{lettrine}
\usepackage{array}
\usepackage{graphicx}
\usepackage{textcomp}
\usepackage{makecell}
\usepackage{graphics} 
\usepackage{epsfig}
\usepackage{tabularray}
\usepackage{subfigure}

\usepackage{epstopdf}
\usepackage{lipsum}

\usepackage{lipsum,amsmath,multicol}
\usepackage{float}
\usepackage{algorithm}
\usepackage{algpseudocode}
\usepackage{wrapfig}
\usepackage{sidecap}

\usepackage{cite}
\usepackage{hyperref}

\usepackage{amsthm}

\usepackage{comment}
\usepackage{array}
\usepackage{glossaries}
\usepackage{breqn}
\makeglossaries
\newcolumntype{L}[1]{>{\raggedright\let\newline\\\arraybackslash\hspace{0pt}}m{#1}}
\newcolumntype{C}[1]{>{\centering\let\newline\\\arraybackslash\hspace{0pt}}m{#1}}
\newcolumntype{R}[1]{>{\raggedleft\let\newline\\\arraybackslash\hspace{0pt}}m{#1}}
\usepackage{sidecap}
\usepackage{url}
\usepackage[T1]{fontenc}
\usepackage{footnote}
\makesavenoteenv{algorithm}
\usepackage{soul}

\usepackage{tabularx}
\usepackage{multicol, blindtext}
\usepackage{multirow}
\usepackage{xcolor}
\usepackage{graphicx}
\usepackage{breqn}

\allowdisplaybreaks

\allowdisplaybreaks
\definecolor{grey}{rgb}{0.5,0.5,0.5}

\newcommand{\polardir}{\boldsymbol{\omega}}

\makeatother
\begin{document} 

%\title{ 
%AMSwarmX: An \textbf{A}lternating 
%\textbf{M}inimization Approach for Safe %Motion Planning of Quadrotor \textbf{Swarm}s %in Comple\textbf{X} Environments}
\title{AMSwarmX: Safe Swarm Coordination in CompleX Environments via Implicit Non-Convex Decomposition of the Obstacle-Free Space}
\author{Vivek K. Adajania, Siqi Zhou, Arun Kumar Singh, and Angela P. Schoellig\thanks{Vivek K. Adajania, Siqi Zhou, and Angela P.~Schoellig are with the Learning Systems and Robotics Lab (http://www.learnsyslab.org) at the University of Toronto Institute for Aerospace Studies, Canada, and the Technical University of Munich, Germany. They are also with the Vector Institute for Artificial Intelligence. Arun Kumar Singh is with the University of Tartu, Estonia.
Emails:
\{vivek.adajania,~siqi.zhou\}@robotics.utias.utoronto.ca,  arun.singh@ut.ee, and angela.schoellig@tum.de.  } 
}
\maketitle

%%%%%%%%%%%%%%%%%%%%%%%%%%% ABSTRACT %%%%%%%%%%%%%%%%%%%%%%%%%%%
\begin{abstract} 
Quadrotor motion planning in complex environments leverage the concept of safe flight corridor (SFC) to facilitate static obstacle avoidance. Typically, SFCs are constructed through convex decomposition of the environment's free space into cuboids, convex polyhedra, or spheres. However, when dealing with a quadrotor swarm, such SFCs can be overly conservative, substantially limiting the available free space for quadrotors to coordinate. This paper presents an Alternating Minimization-based approach that does not require building a conservative free-space approximation. Instead, both static and dynamic collision constraints are treated in a unified manner. Dynamic collisions are handled based on shared position trajectories of the quadrotors. Static obstacle avoidance is coupled with distance queries from the Octomap, providing an implicit non-convex decomposition of free space. As a result, our approach is scalable to arbitrary complex environments. Through extensive comparisons in simulation, we demonstrate a $60\%$ improvement in success rate, an average $1.8\times$ reduction in mission completion time, and an average $23\times$ reduction in per-agent computation time compared to SFC-based approaches. We also experimentally validated our approach using a Crazyflie quadrotor swarm of up to 12 quadrotors in obstacle-rich environments. The code, supplementary materials, and videos are released for reference.

% Many works on quadrotor motion planning in complex environments leverage the concept of a safe flight corridor (SFC) to facilitate static obstacle avoidance. Typically, SFCs are constructed through convex decomposition of the environment's free space into cuboids, convex polyhedra, or spheres. However, when dealing with a quadrotor swarm, this convex decomposition approach can be overly conservative, substantially limiting the available free space for quadrotors to coordinate. This paper presents an Alternating Minimization-based approach that does not require building a conservative free-space approximation. Instead, both static and dynamic collision constraints are treated in a unified manner. Dynamic collisions are handled based on shared position trajectories of the quadrotors. Static obstacle avoidance is coupled with distance queries from the Octomap, providing an implicit non-convex decomposition of free space. As a result, the approach ensures scalability to arbitrary complex environments. Through extensive comparisons in simulation, we demonstrate a $60\%$ improvement in success rate, an average $2\times$ reduction in mission completion time, and an average $23\times$ reduction in per-agent computation time compared to SFC-based approaches. We experimentally validated our approach using a Crazyflie quadrotor swarm of up to 12 quadrotors in obstacle-rich environments. The code, supplementary materials, and videos are released for reference.

\end{abstract}
%%%%%%%%%%%%%%%%%%%%%%%%%%% INTRODUCTION %%%%%%%%%%%%%%%%%%%%%%%%%%%
\section{Introduction}
The excellent maneuverability and agility of quadrotors make them very popular for applications such as search and rescue missions \cite{marconi2012sherpa}, environmental mapping and monitoring \cite{schmuck2017multi}, and payload transport \cite{tang2015mixed}. While single quadrotors are impressive, quadrotor swarms offer even greater advantages, including increased flexibility, efficiency, and robustness \cite{chung2018survey}. As the quadrotors operate in a shared space, they must \textit{coordinate} among themselves to resolve conflicts while also \textit{avoiding static obstacles} present in the environment.

Trajectory optimization approaches for coordinating quadrotor swarms use the ellipsoidal collision avoidance constraints \cite{preiss2017downwash} for inter-agent collision avoidance. These approaches can be divided into two categories: centralized and distributed. Centralized approaches \cite{mip_how, augugliaro2012generation, rastgar2021gpu} solve a joint trajectory optimization problem for all the quadrotors. Despite offering an extended solution space, these approaches become computationally intractable when dealing with a larger number of quadrotors.

\begin{figure}[!t]
    \centering
    \includegraphics[width=\columnwidth]{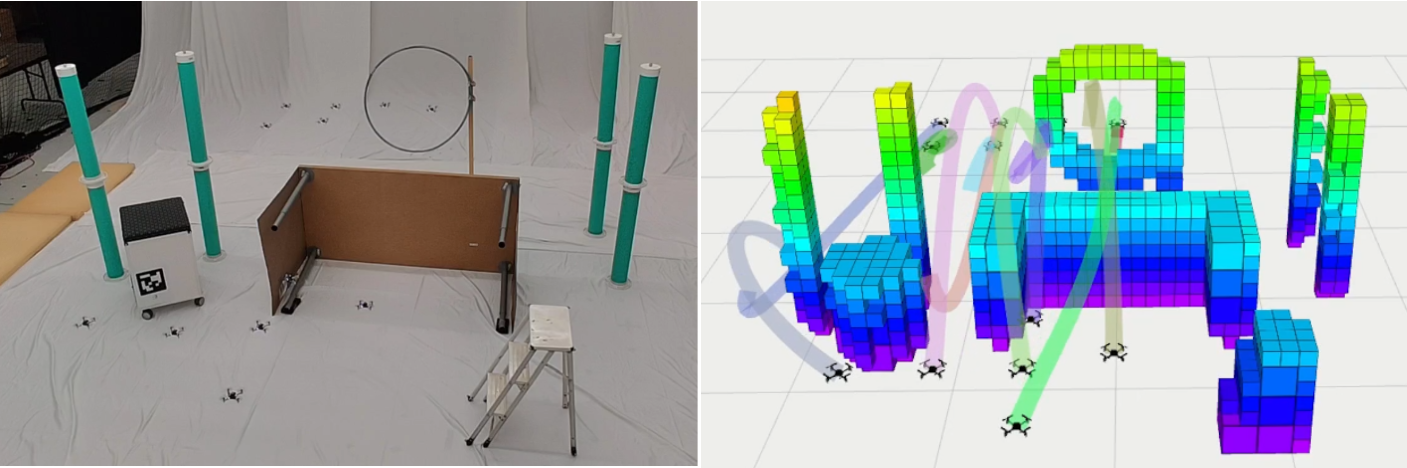}
    \caption{Twelve quadrotors performing a position exchange in a complex environment. Link to video: \protect\hyperlink{http://tiny.cc/AMSwarmXVideo}{\url{http://tiny.cc/AMSwarmXVideo}}. Link to code and supplementary material: \protect\hyperlink{https://github.com/vivek-uka/AMSwarmX}{\url{https://github.com/utiasDSL/AMSwarmX}}.
    }
    \label{teaser}
    \vspace{-0.5cm}
\end{figure}

Distributed approaches such as \cite{luis-ral20, park2022online, amswarm} provide a scalable alternative. In these approaches, each quadrotor independently solves an optimization problem. The collision avoidance constraints are formulated based on the trajectories shared by neighbouring quadrotors. As shown in \cite{amswarm}, the independent optimization problem is a non-convex Quadratically Constrained Quadratic Program (QCQP), arising from non-convex quadratic ellipsoidal collision avoidance and kinematic feasibility constraints. Existing distributed approaches \cite{luis-ral20, park2022online, soria2021distributed} rely on affine approximations: linearizing the collision avoidance constraints and axis-wise decoupling of kinematic constraints. These approximations result in a QP but with small feasible sets. Our previous work \cite{amswarm} showed how to avoid these approximations and still obtained a QP, achieving superior inter-agent collision avoidance performance. 

To navigate a single quadrotor in complex 3D environments, many works have extensively employed the concept of Safe Flight Corridor (SFC) for static obstacle avoidance. Existing works perform convex decomposition of the free space to obtain SFCs, which serve as additional constraint sets in trajectory optimization. Several examples of such convex constraint sets include cuboid \cite{gao2018online}, spheres \cite{ji2021mapless}, and convex polyhedra \cite{liu2017planning}.

SFC-based approaches also utilize high-level path planners such as A* and RRT* to generate a guiding path for trajectory optimization. Some methods \cite{liu2017planning, gao2018online, ji2021mapless} construct a safe corridor around this guiding pah, while others \cite{park2022online} rely on trajectories from previous planning steps. There exists Gradient Descent (GD) based approaches \cite{zhou2020ego, zhou2021ego} that directly incorporate the distance to obstacles, but as a cost, which limits their generalization to different environments.

SFC-based approaches are also popular in both centralized and distributed swarm setups. In \cite{honig2018trajectory, csenbacslar2023rlss}, convex polyhedra are employed for each quadrotor, and authors in \cite{park2020efficient, park2022online} use cuboids to avoid the static obstacles, while inter-agent collision avoidance is formulated via ellipsoidal collision avoidance as previously mentioned. The convex polyhedra and cuboid corridors are represented as affine inequalities and can be incorporated into the QP trajectory generation framework. However, decomposing the free space into these convex regions can be conservative in tight environments, resulting in less free space available for the quadrotors to negotiate around each other.

\paragraph*{Contribution} In this work, we propose an Alternating Minimization (AM) approach that treats static and dynamic collision avoidance constraints in a unified manner yet scales to arbitrary environments. The dynamic inter-agent collisions are handled based on the predicted or shared trajectories of the neighbours while the static obstacle constraints are coupled to the distance queries from the Octomap. We show that our approach implicitly constructs a non-convex decomposition of the free space that is much larger than that of explicit convex decomposition approaches. We compare our approach with state-of-the-art distributed SFC-based approaches from \cite{park2020efficient, amswarm, liu2017planning}. Our simulation comparison shows a 60\% improvement in success rate, on average a 1.8$\times$ reduction in mission completion time, and on average a 23$\times$ reduction in per-agent computation time.

\section{Distributed Motion Planning Problem}
Our objective is to generate smooth, collision-free, and kinematically feasible trajectories that guide $N$ quadrotors from their initial positions $\mathbf{p}_{i,o}$ to their desired goal positions $\mathbf{p}_{i,g}$ within a cluttered and complex environment. The vector $\mathbf{p} = [x, y, z]^T$ represents the three-dimensional position of a quadrotor, with the subscript $i$ denoting the quadrotor index and the subscripts $o$ and $g$ indicating initial and goal variables. Similar to our previous work \cite{amswarm}, we formulate the motion planning for the quadrotor swarm as a distributed trajectory optimization problem. We assume that each quadrotor can communicate with its neighbour without any communication loss or delay. We also assume that each quadrotor has access to a prior map of the environment.

\subsection{Problem Formulation}
\label{problem}
We describe the optimization problem that needs to be solved by quadrotor $i$ at each planning step:
\begin{subequations}
\begin{align}
\min_{\mathbf{p}_i}\hspace{0.5em}& w_{g}\sum_{k=K-\kappa}^{K-1} \left\Vert \mathbf{p}_i[k] - \mathbf{p}_{i,g}\right\Vert^2+w_{s}\sum_{k=0}^{K-1} \left\Vert \mathbf{p}_i^{(q)}[k]\right\Vert^2\label{cost}\\
\text{s.t.}\hspace{0.5em}&     \mathbf{p}_i^{(q)}[0] = \mathbf{p}_{i,a}^{(q)}, \: \forall q =\{0,1,2\} \label{initial_conditions}\\
\hspace{0.5em}&       \left\Vert \dot{\mathbf{p}}_i[k] \right\Vert^2 \leq \overline{v}^2, \:\forall k \label{quad_vel_limits}\\
\hspace{0.5em}&        \underline{f}^2 \leq \left\Vert \ddot{\mathbf{p}}_i[k] + \mathbf{g} \right\Vert^2 \leq \overline{f}^2, \forall k \label{quad_acc_limits} \\
\hspace{0.5em}&           \left\Vert(2s\boldsymbol\Theta_{ij})^{-1}(\mathbf{p}_i[k] - \boldsymbol\xi_j[k])\right\Vert^2 - 1 \geq 0, \:\forall k,j\label{collision_constraint}\\
\hspace{0.5em}& \mathbf{p}_i[k] \in \mathcal{C}_{free}, \:\forall k, \label{static_collision_constraint}
\end{align}
\end{subequations}
where $k$ is the discrete-time index, $K$ is the planning horizon length, $||\cdot||$ denotes the Euclidean norm, and the superscript $(q)$ denotes the $q$-th time derivative of a variable.
The cost function consists of two terms. The first term is the error-to-goal cost applied over the last $\kappa< K$ steps in the prediction horizon; the second term is the smoothness cost that penalizes the $q$-th derivatives of the position trajectory. The constants $w_g$ and $w_s$ are weights of respective terms. 

The equality constraints \eqref{initial_conditions} set the initial position of the trajectory and the higher derivatives to be consistent with the current values of the quadrotor. The inequalities \eqref{quad_vel_limits}-\eqref{quad_acc_limits} enforce bounds on the velocity ($0, \overline{v}$), and the acceleration ($\underline{f}, \overline{f}$). The inequalities \eqref{collision_constraint} enforce the collision avoidance with the $j$-th neighbouring quadrotor with position $\boldsymbol{\xi}_j[k]$. Note that $\boldsymbol{\xi}_j[k]$ is known since the quadrotor communicates the trajectory they computed at the previous planning step to their neighbours. The constant $s$ is the radius of the sphere modelling the quadrotor. $\boldsymbol\Theta_{ij}$ is a diagonal matrix with $(1, 1, 2)$ characterizing an ellipsoidal envelope in the inter-agent collision avoidance. The vector $\mathbf{g} = [ 0,\: 0,\: g]^T$ is the gravitational acceleration vector, where $g$ is the acceleration due to gravity. Constraint \eqref{static_collision_constraint} enforces the quadrotor to remain in the space not occupied by the static obstacles in the environment. 
\subsection{Trajectory Parameterization}
We parameterize the $x$-, $y$-, and $z$-position trajectories for each quadrotor as Bernstein polynomials of degree~$n$. For instance, the $x$-position trajectory for the $i$-th quadrotor is% defined as
\begin{align}
    \begin{bmatrix}x_i[0]&x_i[1]&\ldots&x_i[K-1]
    \end{bmatrix}^T = \mathbf{W}\mathbf{c}_{i,x},
    \label{param}
\end{align}
\normalsize
where $\mathbf{W}\in \mathbb{R}^{K\times(n+1)}$ is the Bernstein basis matrix and $\mathbf{c}_{i,x}$ are the coefficients associated with it. The higher derivatives of the position trajectory have the general form  $\mathbf{W}^{(q)}\mathbf{c}_{i,x}$, where $\mathbf{W}^{(q)}$ is the $q$-th derivative of the Bernstein basis matrix.

\section{Main Algorithmic Results}

This section presents our main algorithmic results. We first describe our novel static obstacle avoidance model and its integration into the AM approach. We then discuss how a discrete path planner can be leveraged to obtain some of the hyperparameters of our trajectory optimizer.

\subsection{Static Obstacle Avoidance Constraints}

\begin{figure}[!t]
    \centering
    \includegraphics[width=0.7\columnwidth]{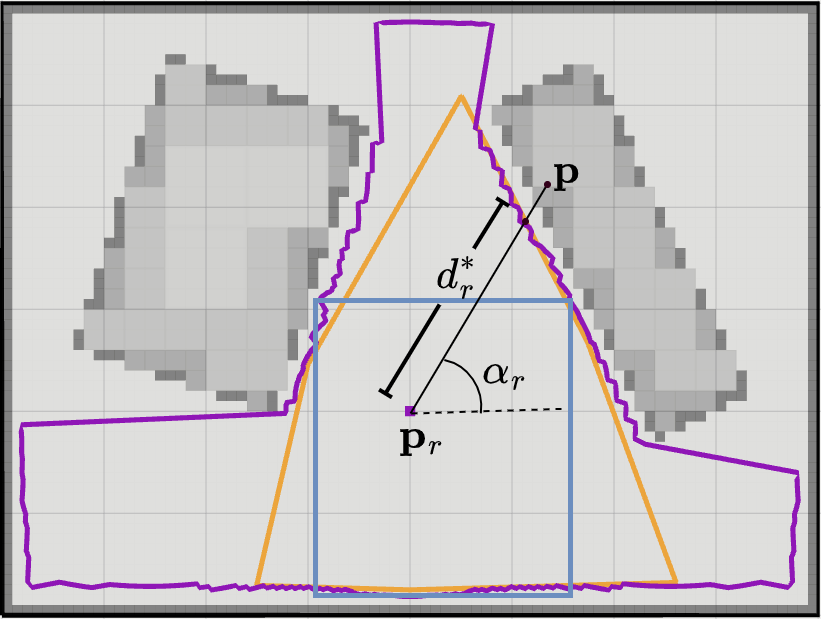}
    \caption{The figure shows the non-convex obstacle-free region (purple) obtained by Octomap distance queries (ray-casting) from a known obstacle-free position $\textbf{p}_r$. Our AM-based approach provides a tractable way of incorporating the purple region into the trajectory optimization. Existing SFC-based approaches decompose the free space into conservative convex sets such as the rectangle \cite{park2022online} (blue) and convex polygon \cite{liu2017planning} (orange).}
    \label{decompose}
    \vspace{-0.5cm}
\end{figure}

Referring to Fig. \ref{decompose}, let $\mathbf{p}_r$ denote a known position in obstacle-free space, and $\mathbf{p}$ represent an arbitrary position in 2D space. We will discuss the possible choices for these positions later in this section. Nevertheless, the condition for $\mathbf{p}$ to be obstacle-free can be expressed as follows:
\begin{align}
    \Vert\mathbf{p} - \mathbf{p}_r\Vert^2 \leq (d_r^*(\alpha_r(\mathbf{p})))^2, 
    \label{coll_free_point}   
\end{align}
where $d_r^*(\alpha_r(\mathbf{p}))$ represents the obstacle clearance from $\mathbf{p}_r$ in the direction of $\alpha_{r}$. We refer to $d_r^*$ as directional clearance and $\mathbf{p}_r$ as the attractor position. Notably, the directional clearance depends on $\alpha_{r}$, which, in turn, relies on $\mathbf{p}$. Moreover, it can be easily obtained through ray-casting on an Octomap while accounting for the dimension of the quadrotor.

As shown in Fig. \ref{decompose}, the constraint \eqref{coll_free_point} implicitly characterizes a non-convex obstacle-free space (shown in purple) in the vicinity of $\mathbf{p}_r$. Existing works often involve computing an explicit convex decomposition of this space, referred to as SFC. For example, two such decompositions in the form of an axis-aligned rectangle and convex polygon are shown in Fig. \ref{decompose}. Clearly, such decompositions are overly conservative, covering only a fraction of the actual obstacle-free space. This severely limits the range of feasible motions for the quadrotor swarms. 

We can extend \eqref{coll_free_point} to 3D by formulating it for quadrotor $i$ at time step $k$ as follows:
\begin{align}
    \Vert\mathbf{p}_i[k] - \mathbf{p}_{i,r}\Vert^2 \leq (d_{i,r}^*(\alpha_{i,r}(\mathbf{p}_i), \beta_{i,r}(\mathbf{p}_i))[k])^2,\: \forall k.\label{obs_avoid_constraint}
\end{align}
Here, $\mathbf{p}_{i,r}$ represents the attractor position associated with quadrotor $i$. The scalar $d_{i,r}^*(\alpha_{i,r}(\mathbf{p}_i), \beta_{i,r}(\mathbf{p}_i))[k]$ denotes the directional clearance from the attractor position in directions $(\alpha_{i,r}[k], \beta_{i,r}[k])$ at each prediction step $k$. We drop the parenthesis and refer to directional clearance as ${d}_{i,r}^*[k]$.

Incorporating constraints of the form \eqref{obs_avoid_constraint} into the optimization problem poses a challenge due to the absence of an analytical, functional form for the directional clearance $d_{i,r}^*$. The subsequent subsections elaborate on how our AM-based trajectory optimizer provides an effective workaround. The key intuition has two core components. First, we initially treat $\alpha_{i, r}[k], \beta_{i, r}[k]$ as independent of $\mathbf{p}_i[k]$ and then gradually enforce their dependency as the optimizer iterations progress (see \eqref{d_obs}). Second, at every step of our AM, we fix  $d_{i,r}^*[k]$ based on some guess of $\alpha_{i, r}[k], \beta_{i, r}[k]$ and we gradually update these guess across iterations. Moreover, we leverage the fact that \eqref{obs_avoid_constraint} becomes a convex quadratic constraint if we fix $d_{i,r}^*[k]$ on the right-hand side.

% $\alpha_{i, r}[k], \beta_{i, r}[k]$, \eqref{obs_avoid_constraint} becomes a quadratic constraint with a constant $d_{i,r}^*[k]$ on the right-hand side.

\subsection{Constraint Reformulation}

The static obstacle avoidance constraints \eqref{obs_avoid_constraint}, inter-agent constraints \eqref{collision_constraint}, acceleration constraints \eqref{quad_acc_limits}, and velocity constraints \eqref{quad_vel_limits} are inherently quadratic in nature. Solving trajectory optimization with these constraints necessitates tackling expensive QCQPs. In this subsection, we undertake the task of reformulating these constraints into a polar form \cite{amswarm}, ultimately enabling us to achieve a QP structure without the need for linearization. We express all these constraints as distinct sets:
\begin{align}
 \mathcal{C}_{i,v}[k] &= \{\dot{\mathbf{p}}_i[k] \in \mathbb{R}^3 \:|\: \mathbf{f}_{i,v}[k]=0,  d_{i,v}[k] \leq \overline{v} \label{d_vel}\},\:\forall k,\\
 \mathcal{C}_{i,a}[k] &= \{\ddot{\mathbf{p}}_i[k] \in \mathbb{R}^3 \:|\: \mathbf{f}_{i,a}[k]=0,  \underline{f} \leq d_{i,a}[k] \leq \overline{f} \label{d_acc}\},\:\forall k,\\
 \mathcal{C}_{ij,c}[k] &= \{\mathbf{p}_i[k] \in \mathbb{R}^3\:|\: \mathbf{f}_{ij,c}[k]=0, d_{ij,c}[k] \geq 1 \label{d_coll}\},\:  \forall k,j,\\
 \mathcal{C}_{i,r}[k] &= \{\mathbf{p}_i[k] \in \mathbb{R}^3\:|\: \mathbf{f}_{i,r}[k]=0, d_{i,r}[k] \leq d^*_{i,r}[k] \label{d_obs}\},\:  \forall k.
\end{align}
In these sets, we introduce functions $\mathbf{f}_{i,v}$, $\mathbf{f}_{i,a}$, $\mathbf{f}_{ij,c}$, and $\mathbf{f}_{i,r}$ defined as follows:
\begin{align}
&\mathbf{f}_{i,v}[k] = \dot{\mathbf{p}}_i[k] - d_{i,v}[k] \cdot \polardir(\alpha_{i,v}[k], \beta_{i,v}[k]),\notag\\
&\mathbf{f}_{i,a}[k] = \ddot{\mathbf{p}}_i[k] + \mathbf{g} - d_{i,a}[k] \cdot \polardir(\alpha_{i,a}[k], \beta_{i,a}[k]),\notag\\
&\mathbf{f}_{ij,c}[k] = (2s\boldsymbol{\Theta}_{ij})^{-1} (\mathbf{p}_i[k] - \boldsymbol\xi_j[k]) \notag\\& \quad \quad \quad \quad - d_{ij,c}[k] \cdot \polardir(\alpha_{ij,c}[k], \beta_{ij,c}[k]),\notag\\
&\mathbf{f}_{i,r}[k] = \mathbf{p}_i[k] - \mathbf{p}_{i,r} - d_{i,r}[k] \cdot \polardir(\alpha_{i,r}[k], \beta_{i,r}[k]), \notag\\
&\polardir(\alpha_{(.)},\beta_{(.)})=[\cos\alpha_{(.)}\sin\beta_{(.)},\sin\alpha_{(.)}\sin\beta_{(.)},\cos\beta_{(.)}]^T \notag.
\end{align}
 Notably, the parameters $(\alpha_{(.)}, \beta_{(.)}, d_{(.)} ) $ represent the polar form representations of the constraints and will be computed by our optimizer concurrently with the trajectory \cite{amswarm}.

It is worth pointing out that our reformulation of inter-gent collision and static obstacle avoidance given by $\mathbf{f}_{ij,c}$ and $\mathbf{f}_{i,r}$ respectively have the same structure. The only difference stems from the fact that for the former, the feasible space of $d_{ij,c}[k]$ is completely defined analytically. In contrast, evaluating the feasibility of $d_{i,r}[k]$ requires distance queries from the Octomap.

\begin{figure*}[!t]
    \centering
    \includegraphics[width=1.6\columnwidth]{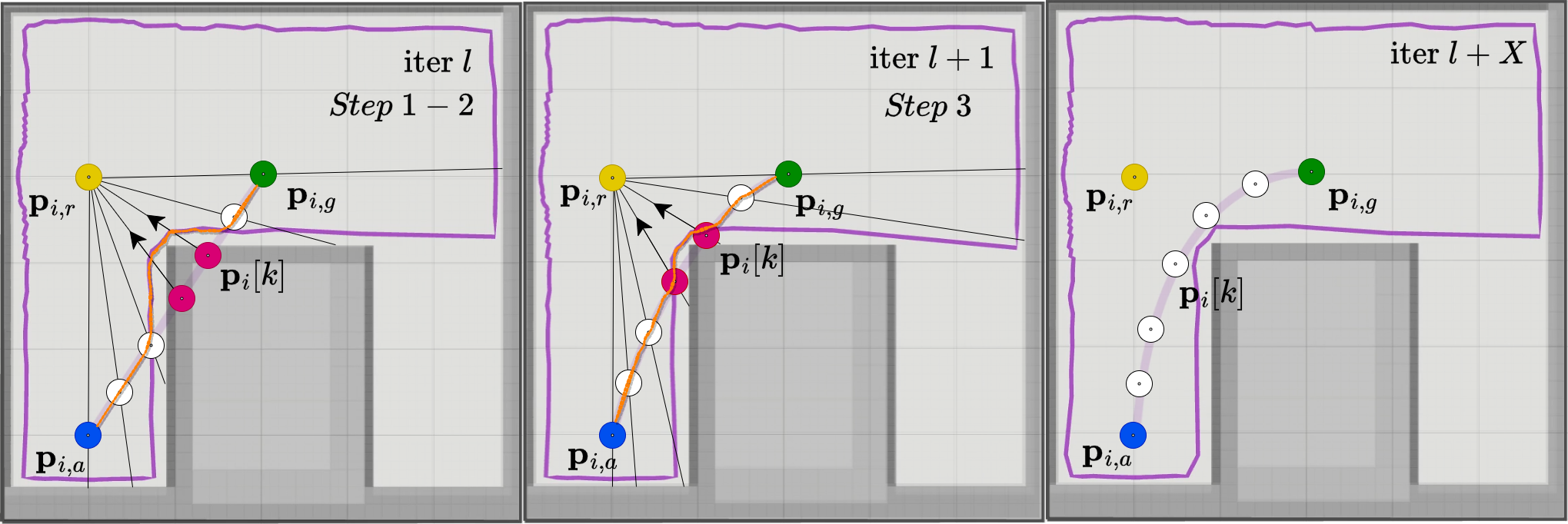}
    \caption{Graphical description of the different steps of our AM-based optimizer. The purple line shows the obstacle-free space estimated by a 360-degree Octomap distance query from the attractor position $\textbf{p}_{i,r}$. Steps 1 and 2 (left) jointly define an approximation (orange line) of the boundary of the obstacle-free boundary. In step 3, the colliding positions (red) are pushed towards the free space using the boundary estimated in steps 1-2. The updated trajectory of step 3 is used to further refine the free-space boundary estimate in the subsequent iterations. The rightmost figure shows the final output from our optimizer.}
    \label{localshape}
    \vspace{-0.5cm}
\end{figure*}
\raggedbottom

\subsection{Reformulated Problem}
We can express the cost function \eqref{cost}, initial conditions \eqref{initial_conditions}, and the polar constraints \eqref{d_vel}-\eqref{d_obs} as a concise optimization problem:
\vspace{-0.1em}
\begin{subequations}
\begin{align}
\min_{\boldsymbol{\zeta}_{i,1}, \boldsymbol{\zeta}_{i,2}, \boldsymbol{\zeta}_{i,3}} \quad & \frac{1}{2} \boldsymbol{\zeta}_{i,1}^T \mathbf{Q} \boldsymbol{\zeta}_{i,1} + \mathbf{q}^T \boldsymbol{\zeta}_{i,1} \label{cost_matrix}\\
\text{s.t.} \quad & \mathbf{A} \boldsymbol{\zeta}_{i,1} = \mathbf{b}(\boldsymbol{\zeta}_{i,2}, \boldsymbol{\zeta}_{i,3}) \label{eq_matrix}\\
& \boldsymbol{\zeta}_{i,1} \in \mathcal{C}_{\boldsymbol{\zeta}_{i,1}}, \quad \boldsymbol{\zeta}_{i,3} \in \mathcal{C}_{\boldsymbol{\zeta}_{i,3}} \label{initial_condi_matrix}
\end{align}
\end{subequations}
Here, $\boldsymbol{\zeta}_{i,1} = [\mathbf{c}_{i,x}^T, \:\mathbf{c}_{i,y}^T,\:\mathbf{c}_{i,z}^T]^T$, $\boldsymbol{\zeta}_{i,2} = [\boldsymbol{\alpha}_{i,c}^T,\: \boldsymbol{\alpha}_{i,a}^T,\: \boldsymbol{\alpha}_{i,v}^T, \: \boldsymbol{\alpha}_{i,r}^T, \:  \boldsymbol{\beta}_{i,c}^T, \: \boldsymbol{\beta}_{i,a}^T, \:  \boldsymbol{\beta}_{i,v}^T,\: \boldsymbol{\beta}_{i,r}^T]^T$, and $\boldsymbol{\zeta}_{i,3} = [\mathbf{d}_{i,c},\: \mathbf{d}_{i,a},\: \mathbf{d}_{i,v}, \mathbf{d}_{i,r}]^T$ represent the optimization variables. Note that the different $\boldsymbol{\alpha}$, $\boldsymbol{\beta}$ and $\mathbf{d}$ are formed by stacking the respective variables at different time steps. For, example $\boldsymbol{\alpha}_{i, r}$ is formed by stacking $\alpha_{i, r}[k]$ for different $k$. The matrix $\mathbf{A}$ and vector $\mathbf{b}$ arise from the equality constraints in the polar form. The matrix $\mathbf{Q}$ and vector $\mathbf{q}$ pertain to the cost function. The sets $\mathcal{C}_{\boldsymbol{\zeta}_{i,1}}$ and $\mathcal{C}_{\boldsymbol{\zeta}_{i,3}}$ correspond to the initial boundary conditions and the inequality constraints from the polar form, respectively. Please refer to Section III-B of \cite{amswarm} for details.

\subsection{Relaxation and Solution using AM}
Our solution process consists of two core steps. First, we relax the equality constraints in the reformulated problem by incorporating them as penalties into the cost function as follows:
\begin{align}
    \min_{\boldsymbol{\zeta}_{i,1} \in \mathcal{C}_{{\zeta}_{i,1}}, \boldsymbol{\zeta}_{i,3} \in \mathcal{C}_{{\zeta}_{i,3}}} \frac{1}{2} \boldsymbol{\zeta}_{i,1}^T\mathbf{Q}\boldsymbol{\zeta}_{i,1} + \mathbf{q}^T\boldsymbol{\zeta}_{i,1} - \langle \boldsymbol\lambda_i, \boldsymbol{\zeta}_{i,1}\rangle \notag \\
    + \frac{\rho}{2}\left\Vert \mathbf{A} \boldsymbol{\zeta}_{i,1} - \mathbf{b}(\boldsymbol{\zeta}_{i,2}, \boldsymbol{\zeta}_{i,3}) \right\Vert^2. \label{augmented_problem}
\end{align}

Here, the penalty parameter $\rho$ and the Lagrange multiplier $\boldsymbol{\lambda}_i$ control the residual of the equality constraints. Next, we apply the AM technique to the relaxed problem. In the following, left superscript $l$ is used to track the values of the variable across iterations. That is, ${^l}(.)$ represents the value of $(.)$ at iteration $l$.

\paragraph*{Step 1} At iteration $l+1$, we fix ${^l}\boldsymbol{\zeta}_{i, 1}$ and ${^l}\boldsymbol{\zeta}_{i, 3}$ to the values obtained at iteration $l$ and only optimize over $\boldsymbol{\zeta}_{i, 2}$. Individual optimizations over different $\boldsymbol{\alpha}, \boldsymbol{\beta}$ in $\boldsymbol\zeta_{i,2}$ can be decoupled into parallel problems with similar structures. For example, the optimization problem for $(\boldsymbol{\alpha}_{i,r}, \boldsymbol{\beta}_{i,r})$ can be reduced to (while ignoring $d_{i, r}^*$ dependency for now):
\vspace{-0.075em}
\begin{align}
    {^{l+1}}\boldsymbol{\alpha}_{i,r}, {^{l+1}}\boldsymbol{\beta}_{i,r} \mathord{=} \arg \min_{\boldsymbol\alpha_{i,r}, \boldsymbol\beta_{i,r}} \notag \\
    \left\Vert  \mathbf{W}{^{l}}\mathbf{c}_{i,x} - \mathbf{x}_{i,r} - {^{l}}\mathbf{d}_{i,r}\cos\boldsymbol{{\alpha}}_{i,r}\sin\boldsymbol{{\beta}}_{i,r} \right\Vert^2 \notag \\
    + \left\Vert \mathbf{W}{^{l}}\mathbf{c}_{i,y} -
    \mathbf{y}_{i,r} - {^{l}}\mathbf{d}_{i,r}\sin\boldsymbol{{\alpha}}_{i,r}\sin\boldsymbol{{\beta}}_{i,r} \right\Vert^2 \notag \\
    + \left\Vert  \mathbf{W}{^{l}}\mathbf{c}_{i,z}-
    \mathbf{z}_{i,r} - {^{l}}\mathbf{d}_{i,r}\cos\boldsymbol{{\beta}}_{i,r} \right\Vert^2.
    \label{alpha_beta_coll}
\end{align}

The solution to \eqref{alpha_beta_coll} can be derived in closed form based on purely geometric considerations \cite{amswarm}. As shown in Fig. \ref{localshape} (left), this step can be geometrically interpreted as obtaining the Octomap distance query directions (black lines) from the attractor position.

\paragraph*{Step 2} In this step, we solve for ${^{l+1}}\boldsymbol\zeta_{i,3}$ while using the known values for ${^{l}}\boldsymbol\zeta_{i,1}, {^{l+1}}\boldsymbol\zeta_{i,2}$. Thus, this step reduces to decoupled problems over $\mathbf{d}_{i, c}, \mathbf{d}_{i, r}, \mathbf{d}_{i, v}, \mathbf{d}_{i, a}$. For example, the optimization over $\mathbf{d}_{i, r}$ can be expressed as:
\vspace{-0.075em}
\begin{align}
    {^{l+1}}\mathbf{d}_{i,r} = \arg \min_{\mathbf{d}_{i,r} \leq \mathbf{d}_{i,r}^*} \notag \\
    \left\Vert  \mathbf{W}{^{l+1}}\mathbf{c}_{i,x} - \mathbf{x}_{i,r} - \mathbf{d}_{i,r}\cos{^{l+1}}\boldsymbol{{\alpha}}_{i,r}\sin{^{l+1}}\boldsymbol{{\beta}}_{i,r} \right\Vert^2 \notag \\
    + \left\Vert \mathbf{W}{^{l+1}}\mathbf{c}_{i,y} -
    \mathbf{y}_{i,r} - \mathbf{d}_{i,r}\sin{^{l+1}}\boldsymbol{{\alpha}}_{i,r}\sin{^{l+1}}\boldsymbol{{\beta}}_{i,r} \right\Vert^2 \notag \\
    + \left\Vert  \mathbf{W}{^{l+1}}\mathbf{c}_{i,z}-
    \mathbf{z}_{i,r} - \mathbf{d}_{i,r}\cos{^{l+1}}\boldsymbol{{\beta}}_{i,r} \right\Vert^2.
    \label{d_coll_sol}
\end{align}
Since ${^{l+1}}\boldsymbol{\alpha}_{i, r}, {^{l+1}}\boldsymbol{\beta}_{i, r}$ have already been obtained in the previous step, we can use them to determine the directional clearance $\mathbf{d}_{i, r}^*$ from the Octomap queries for each prediction step $k$. Thus, \eqref{d_coll_sol} reduces to a QP with a closed form solution \cite{amswarm}.

\paragraph*{Step 3} In this step, we use the known values of  ${^{l+1}}\boldsymbol{\zeta}_{i,2}, {^{l+1}}\boldsymbol{\zeta}_{i,3}$ to optimize over just $\boldsymbol\zeta_{i,1}$. This reduction has two important implications. First, fixing   $\boldsymbol{\alpha}_{i, r}$ (from $\boldsymbol{\zeta}_{i,2}$ ) and $\mathbf{d}_{i, r}$ (from $\boldsymbol{\zeta}_{i,3}$ ) allows us to construct an estimate of the boundary of the obstacle-free space (orange strip in Fig. \ref{localshape} (left)) as $\mathbf{p}_{i,r}[k] +d_{i,r}[k] \cdot \polardir[k]$ (recall \eqref{d_obs}). Second, \eqref{augmented_problem} is transformed into an equality-constrained QP with a closed-form solution. Moreover,  geometrically, the effect of this QP is to push the trajectory positions in collision with the obstacle towards the boundary of the obstacle-free space (Fig. \ref{localshape} (left)). The solution of this step will be fed to the next iteration. It will lead to further refinement of $\boldsymbol{\zeta}_{i,2}, \boldsymbol{\zeta}_{i,3}$ and consequently an updated definition of the boundary of the obstacle-free space (Fig.\ref{localshape} (middle)).

A few additional points about our AM-based approach are worth pointing out. As shown in Fig.\ref{localshape}, the description of obstacle-free space (orange strip) as used by our optimizer is slightly conservative when compared to the true boundary (shown in purple). But importantly, by construction the estimated boundary overlaps with the true boundary for the colliding trajectory segment.

\paragraph*{Step 4} The Lagrange multiplier $\boldsymbol\lambda_i$ is updated using the gradient of the penalty term \cite{admm_neural}. We increment the penalty parameter $\rho$ by $\Delta \rho$ and repeat Steps 1 to 4 until the residuals of the penalty term fall below a predefined threshold. A typical final output is presented in Fig. \ref{localshape} (right). We recommend watching the video at the following link: \url{http://tiny.cc/AMIterViz}.

\subsection{Visibility Condition}
\begin{figure}[!t]
    \vspace{0.05em}
    \centering
    \includegraphics[width=0.8\columnwidth]{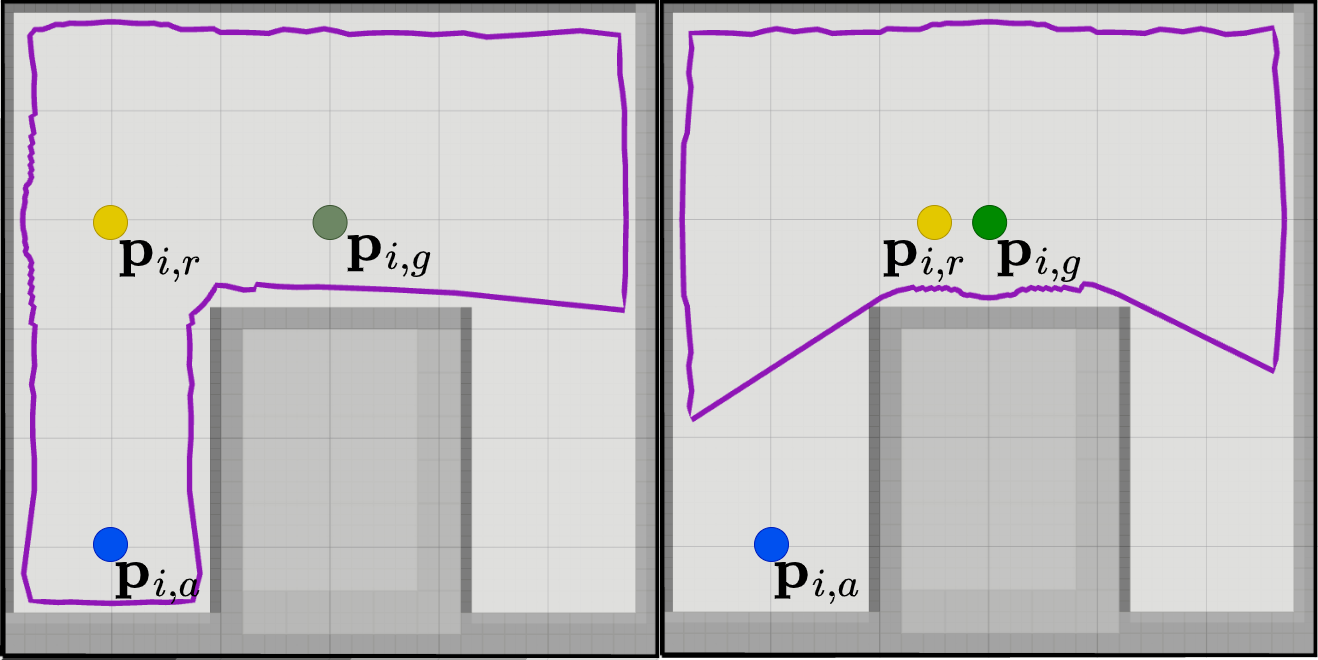}
    \caption{The visibility condition says that the attractor position $\mathbf{p}_{i,r}$ should be visible from the current $\mathbf{p}_{i,a}$ and the goal position $\mathbf{p}_{i,g}$, and vice-versa. In the left figure, the condition is met indicating all three of them lie in the same feasible space of the static obstacle avoidance constraints. While in the right figure, the condition is not met and the AM algorithm may not find an obstacle-free solution.}
    \label{condition}
    \vspace{-0.5cm}
\end{figure}

The performance of our AM optimizer depends on the position of the attractor $\mathbf{p}_{i, r}$. Empirically, we have observed the best performance when $\mathbf{p}_{i, r}$ is visible from both the current position ($\mathbf{p}_{i, a}$) and the goal position ($\mathbf{p}_{i, g}$), and vice versa. The intuition behind this condition is visualized in Fig. \ref{condition}. In the left figure of Fig. \ref{condition}, the visibility condition is met, ensuring that $\mathbf{p}_{i, r}$, $\mathbf{p}_{i, a}$, and $\mathbf{p}_{i, g}$ all lie within the obstacle-free space constructed around $\mathbf{p}_{i, r}$ using Octomap distance queries. In contrast, when the visibility condition is not met (for example, when the current position is outside the constructed obstacle-free space, as shown in the right figure), using the AM optimizer may lead to infeasible solutions. Similarly, if the goal position lies outside the feasible obstacle-free space, the quadrotor will not make any progress toward it, and the generated trajectory will remain confined to the feasible obstacle-free space. We ensure the visibility condition by carefully selecting $\mathbf{p}_{i, r}$ and incorporating an intermediate goal selection routine within our pipeline, which is described in the following section.

\raggedbottom

\subsection{Discrete Path Planning}
We employ an off-the-shelf discrete path planner, A*, on the prior map and use a simple heuristic for meeting the visibility condition. This path planner generates an array $\mathbf{q}_{i,\text{gp}}$ consisting of obstacle-free positions that connect the current position to the final goal position. From this array, we follow a two-step process. We first select the last visible position from the current position within the array. This selected position becomes our attractor. Next, starting from the obtained attractor position, we choose the last visible position as our intermediate goal. As the quadrotor moves, we select a new attractor and intermediate goal at each planning step. Eventually, the intermediate goal would converge to the final goal position. Note that we account for the dimension of the quadrotor when checking for visibility.

\subsection{Summary of Proposed Approach}
 Algorithm \ref{overall_approach} describes all the components $i^{th}$ quadrotor would use to navigate in a complex 3D environment. The input to the algorithm is the current state, desired goal position and map information. First, the quadrotor $i$ receives planned trajectories from neighbouring quadrotors, allowing inter-agent collision avoidance constraints to be formulated. Second, it runs A* on the prior map to obtain a path. Third, from this path, an attractor and an intermediate goal position satisfying visibility conditions are selected for the formulation of static obstacle avoidance constraints. Finally, the quadrotor builds the reformulated problem and applies the AM optimizer to generate the trajectory. These steps are repeated in the next planning step.

\begin{algorithm}[!t]
  {\footnotesize
  \textbf{Input} Current state $\mathbf{p}_{i,a}^{(q)}$, Final goal $\mathbf{p}_{i,g}$, Octomap
  $\mathcal{W}$\\
  \textbf{Output} Trajectory coefficients $\boldsymbol{\zeta}_{i,1}$
 \caption{\texttt{generateTrajectory}}    
    \begin{algorithmic}[1]
    \State $\boldsymbol{\xi}_j \leftarrow$ \texttt{NeighbouringQuadrotorsTrajectories} 
    \State $\mathbf{q}_{i,gp}$ $\leftarrow$ \texttt{runGridPlanner}$(\mathbf{p}_{i,a}, \mathbf{p}_{i,g}, \mathcal{W})$
    \State $\mathbf{p}_{i,r}$ $\leftarrow$ \texttt{selectAttractorPosition}$(\mathbf{q}_{i,gp}, \mathbf{p}_{i,a})$
    \State $\mathbf{p}_{i,w}$ $\leftarrow$ \texttt{selectGoalPosition}$(\mathbf{q}_{i,gp},$  $\mathbf{p}_{i,r})$
    \State \texttt{buildReformulatedProblem}$(\mathbf{p}_{i,a}, \mathbf{p}_{i,w}, \mathbf{p}_{i,r}, \boldsymbol{\xi}_j)$
    \State $\boldsymbol{\zeta}_{i,1} \leftarrow$  \texttt{alternatingMinimization}$(\mathcal{W})$
    \State \textbf{return} $\boldsymbol{\zeta}_{i,1}$
    \end{algorithmic}
    \label{overall_approach}
    \normalsize}
\end{algorithm}

\section{Validation and Benchmarking}
\begin{figure}[!t]
    \centering
    \includegraphics[width=\columnwidth]{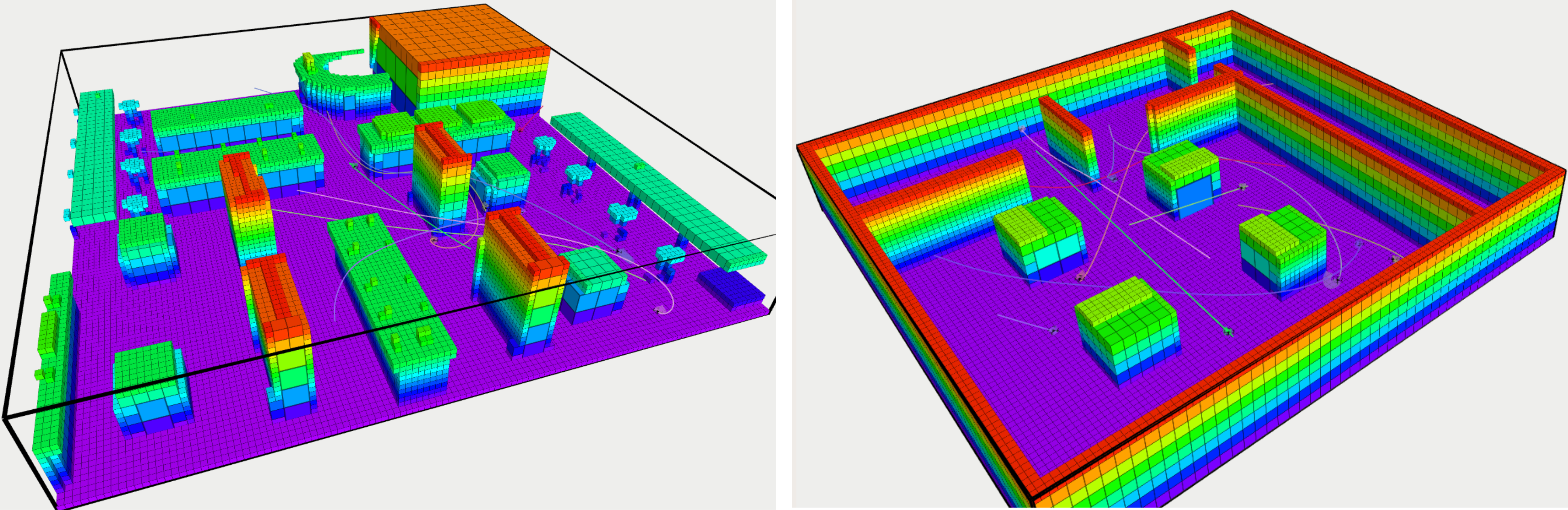}
    \caption{We conduct a total of $60$ simulation runs of different swarm planning approaches in a bookstore (left) and random room (right) environment. The dimension of the environment is $12m\times12m\times2.5m$ for the bookstore and $10m\times10m\times2m$ for the random room.}
    \label{envs}
    \vspace{-0.5cm}
\end{figure}

This section presents a comprehensive comparison in a simulation of our proposed approach with state-of-the-art baselines \cite{amswarm,liu2017planning,park2022online} and experimental validation. All simulations and experiments were executed on a PC with Intel Xeon CPU with 8 cores and 16 GB of RAM, running at 3 GHz. Our simulation evaluation includes two complex environments: the "bookstore" and the "random room" see Fig. \ref{envs}. We conducted a total of $60$ trials, $30$ in each environment, using swarm sizes from $10$ to $50$, and all with randomized start-goal positions. A trial is successful if all quadrotors reach their designated goal positions within a time limit of $60s$ while avoiding collisions. The C++ implementation of the proposed approach, baselines and the parameters used can be found here: \cite{github_amswarm}. We refer to our proposed approach as "AMSwarmX". 

\raggedbottom

\subsection{Distributed Swarm Baselines}
 We compare AMSwarmX with the following two distributed swarm baselines:

 \paragraph*{{LSC-Planner} \cite{park2022online}} {This approach uses Octomap for environment representation and subsequent construction of SFC in the form of axis-aligned cuboids \cite{park2020efficient} for static obstacle avoidance. Inter-agent collision avoidance is achieved by incorporating ellipsoidal collision avoidance constraints, which are linearized using the convex hull property of the Bernstein polynomial. It also employs A* as a high-level path planner and chooses a visible position from current position on the planned path as the intermediate goal. The original approach includes deadlock resolution, but we exclude it in our comparison to focus on collision avoidance capabilities. Note that a deadlock resolution strategy can potentially enhance the performance of any approach. With this baseline, we showcase the advantages provided by a unified linearization-free treatment of static and dynamic collision avoidance.}

  \paragraph*{AMSwarmED (combination of \cite{amswarm} and \cite{liu2017planning})}This baseline is a combination of our prior work \cite{amswarm} augmented with the ellipsoidal free space decomposition method to compute a convex polyhedron \cite{liu2017planning}. The use of high-level path planner and intermediate goals is the same as the previous baseline. Since the dynamic inter-agent collision avoidance part of \cite{amswarm} is the same as our current work, this baseline essentially validates the efficacy of our implicit non-convex free space decomposition.

\begin{figure}[!t]
	\centering
	\begin{minipage}{.325\columnwidth}
		\centering
		\includegraphics[width=\textwidth]{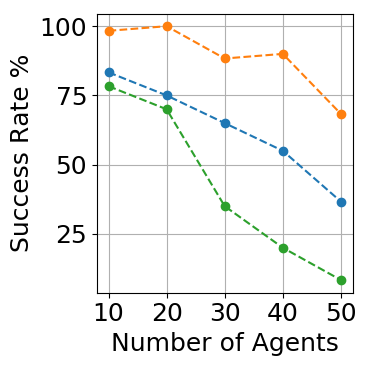}
	\end{minipage}~
	\begin{minipage}{.325\columnwidth}
		\centering
		\includegraphics[width=\textwidth]{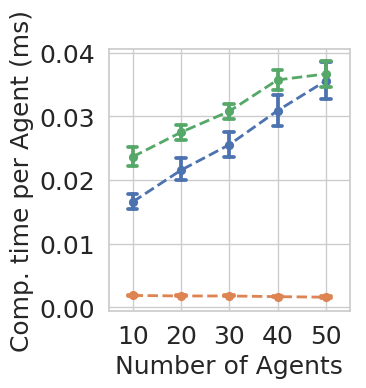}
	\end{minipage}~
	\begin{minipage}{.325\columnwidth}
		\centering
		\includegraphics[width=\textwidth]{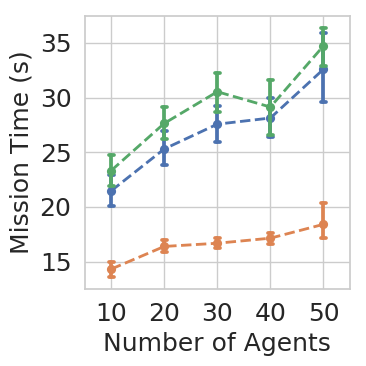}
        \end{minipage}
         \centering
	   \includegraphics[width=0.8\columnwidth]{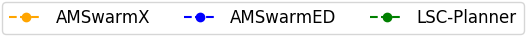}
	\caption{Performance comparison of the different approaches in a point-to-point transition setting with an increasing swarm size. Sixty random trials were run for each swarm size, and the averages are plotted.}
 	\label{comparison_fig}
  \vspace{-0.5cm}
\end{figure}
\raggedbottom

\subsection{Comparative Analysis}
\paragraph*{Success Rate} 
Fig. \ref{comparison_fig} (left) shows the improvement achieved by the AMSwarmX approach over AMSwarmED and LSC-Planner. For swarm sizes up to 20, the success rate of LSC-Planner is similar to that of AMSwarmED, while AMSwarmX outperforms both approaches. However, as the swarm size increases, the performance of LSC-Planner degrades significantly compared to AMSwarmED and AMSwarmX.
As mentioned earlier, LSC-Planner linearizes the inter-agent collision avoidance constraints, leading to hyperplane constraints that are known to be conservative \cite{amswarm}. Also, the cuboid decomposition of obstacle-free space, especially in tight space, is conservative. Consequently, the quadrotors often end up in deadlocks. In contrast, both AMSwarmED and AMSwarmX can directly handle the quadratic form of inter-agent collision avoidance constraints in the optimizer. AMSwarmED performs better than LSC-Planner, as the convex polyhedron corridor provides larger free space for the quadrotors to maneuver. However, AMSwarmX consistently outperforms both approaches, showing an improvement of $15\%$-$60\%$, validating the benefits of our proposed static obstacle avoidance strategy.

\paragraph*{Computation Time}
The middle plot in Fig.~\ref{comparison_fig} shows the computation time per agent for all approaches. There is only a small (max. $20ms$) difference between the mean values for swarm sizes $10$ to $50$. This can be attributed to the distributed nature of the approaches, with each quadrotor solving its own optimization problem and considering only neighbouring quadrotors during optimization. AMSwarmX has the lowest average computation time per quadrotor, showing a 21$\times$ and 23$\times$ reduction compared to AMSwarmED and LSC-Planner, respectively. This is because AMSwarmX adds only one static obstacle avoidance constraint at each planning step. In contrast, AMSwarmED adds numerous hyperplane constraints stemming from the convex polyhedron, and similarly, LSC-Planner adds numerous hyperplane constraints stemming from cuboid corridors. Additionally, LSC-Planner employs numerous polynomial pieces, which increases the number of decision variables. It is important to note that the plot only shows the time required to solve the optimization problem, while the time taken to generate a discrete path and a SFC is in the sub-millisecond range.

\paragraph*{Mission Time}
The rightmost plot in Fig.~\ref{comparison_fig} shows mission completion times. LSC-Planner performs the worst due to non-smooth transitions around corners caused by the cuboid corridors. AMSwarmED, benefiting from a better decomposition, exhibits smoother transitions compared to LSC-Planner. AMSwarmX achieves the smoothest transitions as it better captures the local shape of the free space (see accompanying video). Cuboid-shaped corridors and convex polyhedron are overly conservative in tight spaces, making it difficult for quadrotors to maneuver around each other, resulting in increased mission completion times. Overall, AMSwarmX demonstrates a time reduction of $1.75\times$ and $1.87\times$ over AMSwarmED and LSC-Planner, respectively.

\subsection{Experimental Validation}
We conducted the experimental validation of AMSwarmX using our Crazyflie 2.0 swarm testbed in two complex environments. Trajectories were computed on a single computer, with a CPU thread assigned to each quadrotor. These computed trajectories were transmitted to the lower-level controller at each planning step. We provided AMSwarmX with the Octomap representation. In both scenarios, the quadrotors perform three transitions: first, the quadrotors execute a position exchange; next, a random transition; and finally, they return to their original take-off positions. The average per-agent computation time, inter-agent distance, and distance to obstacles were found to be $4.4ms$, $0.43m$, and $0.37m$, respectively. The smallest inter-agent distance and distance to obstacles were $0.26m$ and $0.1m$, respectively. The quadrotors successfully complete the task without collisions. The demonstration video can be found here: \url{http://tiny.cc/AMSwarmXVideo} and also in the submitted supplementary media file.

\section{Conclusions and Future Work}
In this work, we have taken a step towards deploying quadrotor swarms in complex 3D environments. We have introduced a novel approach that addresses dynamic collisions through shared trajectories without linearization. Meanwhile, static collisions are handled by exploiting Octomap distance queries to build an implicit non-convex decomposition of free space. Our approach allows quadrotors to utilize available free space more efficiently, and thus, we outperform SFC-based methods in terms of mission completion time, success rate, and per-agent computation time. Furthermore, we have conducted experimental validations of our approach using a Crazyflie swarm testbed. Our work assumes prior map information but this requirement can be relaxed. For instance, our pipeline can be integrated with a high-level exploration planner \cite{dang2020graph} that generates paths based on local map information. Our future research efforts are directed towards the development of high-level path planners tailored to specific applications such as warehouse inventory management, surveillance, and exploration.

\bibliography{main}
\bibliographystyle{IEEEtran}
\end{document}